\pdfoutput=1

\documentclass[11pt]{article}

\usepackage[final]{acl}

\usepackage{times}
\usepackage{latexsym}

\usepackage[T1]{fontenc}

\usepackage[utf8]{inputenc}

\usepackage{microtype}

\usepackage{inconsolata}

\usepackage{times}
\usepackage{latexsym}
\usepackage{amssymb}
\usepackage{amsmath}
\usepackage{graphicx}
\usepackage{subcaption}
\usepackage{subfloat}
\usepackage{diagbox}
\usepackage{tikz}
\usepackage{colortbl}
\usepackage{xcolor}
\usepackage{multirow}
\usepackage{caption}
\usepackage{subcaption}
\usepackage{subfloat}
\usepackage{amsmath}
\usepackage{enumerate}
\usepackage{diagbox}
\usepackage{tikz}
\usepackage{colortbl}
\usepackage{xcolor}
\usepackage{multirow}
\usepackage{caption}
\usepackage{subcaption}
\usepackage{subfloat}
\usepackage{amsmath}
\usepackage{enumerate}
\usepackage{stfloats}
\usepackage[linesnumbered,ruled,vlined]{algorithm2e}

\title{Whispers that Shake Foundations: Analyzing and Mitigating False Premise Hallucinations in Large Language Models}

\author{Hongbang Yuan \textsuperscript{1,2}, Pengfei Cao \textsuperscript{1,2},  Zhuoran Jin \textsuperscript{1,2}, Yubo Chen \textsuperscript{1,2} \\ \textbf{Daojian Zeng \textsuperscript{3}, Kang Liu \textsuperscript{1,2}, Jun Zhao \textsuperscript{1,2}} \\ \textsuperscript{1}The Laboratory of Cognition and Decision Intelligence for Complex Systems, \\ Institute of Automation, Chinese Academy of Sciences, Beijing, China  \\ \textsuperscript{2}School of Artificial Intelligence, University of Chinese Academy of Sciences, Beijing, China  \\ \textsuperscript{3}Hunan Normal University, Changsha, China \\
        \footnotesize{\texttt{\{hongbang.yuan, pengfei.cao, zhuoran.jin, yubo.chen, kliu, jzhao\} @nlpr.ia.ac.cn }} }

\begin{document}
\maketitle
\begin{abstract}
Large Language Models (LLMs) have shown impressive capabilities but still suffer from the issue of hallucinations.  A significant type of this issue is the false premise hallucination, which we define as the phenomenon when LLMs generate hallucinated text when confronted with false premise questions. In this paper, we perform a comprehensive analysis of the false premise hallucination and elucidate its internal working mechanism: a small subset of attention heads (which we designate as false premise heads) disturb the knowledge extraction process, leading to the occurrence of false premise hallucination. Based on our analysis, we propose \textbf{FAITH} (\textbf{F}alse premise \textbf{A}ttention head constra\textbf{I}ining for mi\textbf{T}igating \textbf{H}allucinations), a novel and effective method to mitigate false premise hallucinations. It constrains the false premise attention heads during the model inference process. Impressively, extensive experiments demonstrate that constraining only approximately $1\%$ of the attention heads in the model yields a notable increase of nearly $20\%$ of model performance.
\end{abstract}
\section{Introduction}

\begin{figure}
\centering

\includegraphics[width=\linewidth]{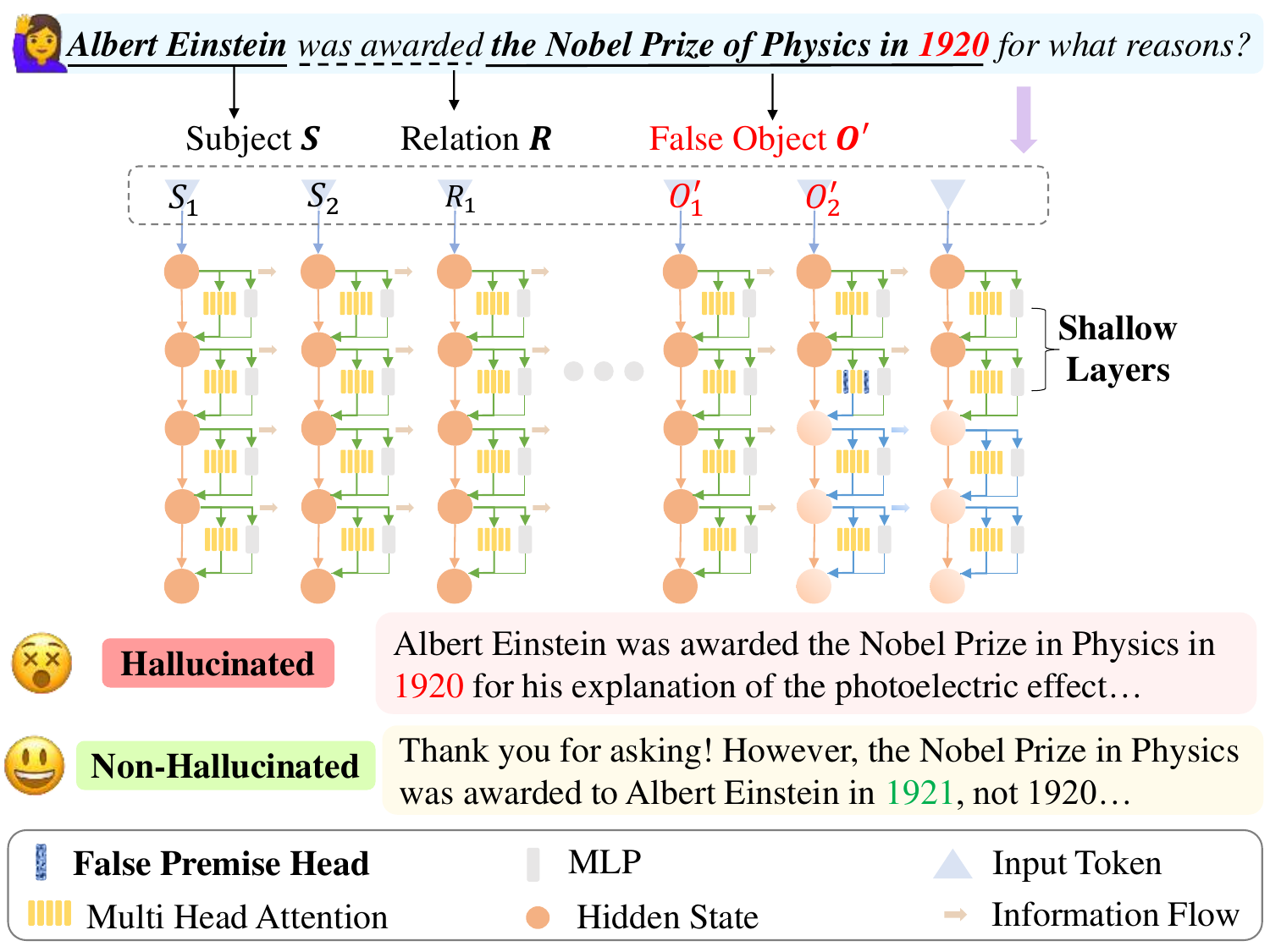}
\caption{Illustration of the false premise hallucination. The question contains the false premise that ``\textit{Albert Einstein was awarded The Nobel Prize of Physics in 1920}'' whereas in fact he was awarded the prize in 1921. We find that the presence of false premise attention heads contributes to the hallucinated response. Our method can effectively mitigate the false premise hallucination.}
\label{introduction}
\end{figure}

Large language models (LLMs) have shown impressive capabilities \citep{wei2023chainofthought, xu2023tool-manipulation-capability} and achieved remarkable success in many tasks \citep{bubeck2023sparks_of_agi,bang2023multitask-chatgpt-evaluation,jiao2023chatgpt-a-good-translator}. However, they often generate texts that are seemingly plausible but deviate from factual knowledge sources \cite{zhang2023sirens_song}, which is a severe problem known as \textbf{hallucination}. 

To address this, many studies focus on detecting \citep{manakul2023selfcheckgpt,min2023factscore} and mitigating hallucinations \citep{trivedi-etal-2023-interleaving-retrieval,gou2023critic}. However, few of them pay attention to a particular type of hallucination: \textbf{False Premise Hallucination}. We define it as the phenomenon in which LLMs generate hallucinated texts in response to the false premise question. False premise questions are questions that contain falsely assumed facts that are not directly stated but are likely to be believed by the asker \citep{yu-etal-2023-crepe,kim-etal-2023-qa}. For these questions,  LLMs tend to respond directly without explicitly verifying their plausibility despite the corresponding factual knowledge can be recalled straightforwardly. For example, as shown in Figure \ref{introduction}, the user query on the top contains a false premise: \textit{(Albert Einstein, was awarded, The Nobel Prize of Physics in 1920)}, denoted as (\textit{subject, relation, false object}). LLMs are able to generate the correct time ``\textit{1921}'' when directly queried about the award time but produce hallucinated time ``\textit{1920}'' in response to the false premise question. 

The exploration of false premise hallucination is exceptionally  significant and valuable. While false premise questions are pervasive on the Internet and users are highly likely to pose these questions when interacting with the LLMs, LLMs are prone to generate hallucinated texts when confronted with such questions. For example, about 25\% of the questions on the discussion website Reddit contain false premises \citep{yu-etal-2023-crepe,fan2019eli5}. According to our statistics, Llama-2-13b achieves an accuracy of $100\%$ on our collected 4004 direct-asking questions but drops to only $24.3\%$ on the corresponding false premise questions. However, the analysis of false premise hallucination is non-trivial. Intuitively, LLMs generate hallucinated texts due to a lack of pertinent factual knowledge \citep{zhang2023sirens_song,huang2023survey_of_hallucination}. But false premise hallucination introduces a more intricate scenario, wherein LLMs still generate hallucinated texts even when the corresponding factual knowledge is already stored in their parameters.

In this paper, we conduct a comprehensive analysis of false premise hallucination and elucidate its internal working mechanism. Firstly, prior to the analysis, we propose an automatic dataset construction pipeline for the evaluation of false premise hallucination and create two representative and easy-to-evaluate datasets based on it. Then, we investigate the external manifestation of false premise hallucination and observe that LLMs exhibit more inherent uncertainty when generating hallucinated answers. Subsequently, to reveal the source of the uncertainty, we delve into the internal information flow during the hallucination occurrences. We discover that knowledge about the subject stored in model parameters is disturbed in the shallow layers of the model, particularly around the false object mentioned in the question. 
Furthermore, as many studies \citep{meng2022locating-and-editing,yuksekgonul2023attention-satisfies} indicate that self-attention layers transfer the factual knowledge stored in the Multi-Layer Perception (MLP) layers, we explore the internals of the self-attention layers and investigate the influence of each individual attention head on specific pieces of factual knowledge. We find out that a small set of attention heads consistently exert great influence on the factual knowledge across almost all the samples and we name them as \textbf{False Premise Heads}. As depicted in Figure \ref{introduction}, the false premise heads predominantly reside in the shallow layers, functioning around the false object mentioned in the question. Experimental results demonstrate that the presence of false premise heads disturb the extraction of the factual knowledge about the subject, leading to false premise hallucinations.

Based on our analysis, we propose a novel method termed \textbf{FAITH} (\textbf{F}alse premise \textbf{A}ttention head constra\textbf{I}ning for mi\textbf{T}igating \textbf{H}allucinations) to  mitigate hallucinations. It localizes the false premise attention heads for a group of false premise questions and subsequently constrains the function of these attention heads during the model inference process.
Extensive experiments demonstrate the effectiveness of our method comparing with the baseline methods.

Our primary contributions can be summarized as follows:

\begin{itemize}

    \item We propose an automatic dataset construction pipeline for the evaluation of false premise hallucination and create two representative and easy-to-evaluate datasets to facilitate analysis.
    \item We conduct an in-depth analysis of the false premise hallucination from the surface to the internals of LLMs and elucidate its internal working mechanism by revealing the presence of false premise attention heads.

    \item We propose FAITH, a novel method to mitigate false premise hallucinations based on our in-depth analysis. 
    Impressively, extensive experiments demonstrate that constraining only approximately $1\%$ of all the attention heads in the model yields a notable increase of nearly $20\%$ in accuracy, which is highly effective. \footnote{Our code and datasets will be available after acceptance.}

\end{itemize}



\section{Background} 
\label{transformer_architecture}
In this section, we briefly describe the transformer architecture \citep{vaswani2017attention-is-all-you-need} in autoregressive, decoder-only language models from the perspective of residual stream \citep{elhage2021mathematical-framework-for-transformer-circuits}.

Given the context $\{t_{1},t_{2},...,t_{N}\}$ consisting of $N$ tokens, 
the transformer architecture starts with a combination of token embeddings and position embeddings $x_{0} \in \mathbb{R}^{N \times d}$, where $d$ is the model dimension. It marks the start of the residual stream, with a series of residual layers that read from the stream and write back their processed results. Each residual layer is comprised of a self-attention layer and a MLP layer. The information update of each residual layer can be expressed as:
\begin{equation*}
    x_{l} = x_{l-1} + a_{l} + m_{l}
\end{equation*}
where $x_{l}$ is the hidden state after the $l$-th layer, $a_{l}$ is the output of the self-attention layer and $m_{l}$ is the output of the MLP layer. More specifically, the calculation of the MLP layer is: 
\begin{equation*}
    m_{l} = f(x_{l-1} K^T)V
\end{equation*}
where $K,V \in \mathbb{R}^{dm\times d}$ and $f$ is a non-linear function. The self-attention contribution is 
\begin{equation*}
\begin{aligned}
        a_{l} &= \sum_{h=1}^{H} A^h (x_{l-1} W^{h}_V) W^{h}_{O} \\
    A^h &= softmax(\frac{(x_{l-1}W^h_Q) (x_{l-1}W_K^h)^T}{\sqrt{d_h/H}})
\end{aligned}
\end{equation*}
where $W_K^h,W^h_Q,W^{h}_{O} \in \mathbb{R}^{d\times d_h},W^{h}_{O} \in \mathbb{R}^{d_h\times d}$ are the parameter matrices, $H$ is the number of attention heads, $d_h=d/H$ is the hidden dimension of each head and
$A \in \mathbb{R}^{N\times N}$  is a lower triangular attention pattern matrix, showing the interaction between tokens in different layers. After $L$ residual layers, a layer norm is applied and then an unembedding matrix $W_U\in \mathbb{R}^{d\times V}$ projects the hidden state $x_{L}$ to logits, where $V$ is the length of the vocabulary. 

\section{Dataset Construction}
\label{dataset_construction}
In this section, we describe our proposed automatic dataset construction pipeline for the evaluation of false premise hallucination and provide the details of our constructed dataset.

To prevent the memorization of the question  \citep{carlini2022quantifying-memorization,ramakrishna-etal-2023-invite-testbed} and facilitate the incorporation of the evolving new knowledge, we propose an automatic dataset construction pipeline, which can be divided into the following three stages:


(1) \textbf{Triple Selection} We select a set of factual triples using WikiData \footnote{https://query.wikidata.org/}. We assess whether the factual triple $(s,r,o)$ is stored in the model parameters by asking a question with only subject $s$ and relation $r$.
We retain the triple only if the object $o$ is present in the answer.
(2) \textbf{Triple Corruption} We replace the object $o$ in the original triple $(s,r,o)$ with an incorrect entity $o'$ to obtain the corrupted triple $(s,r,o')$. For example, we transform the original triple ``\textit{(Albert Einstein, was awarded, the Nobel Prize of Physics in 1921)}'' into the corrupted triple ``\textit{(Albert Einstein, was awarded, the Nobel Prize of Physics in 1920)}''.
(3) \textbf{Question Construction} We construct a false premise question by filling a predefined question template with the previous corrupted triple $(s,r,o')$. For example, we define one of the question templates as ``\textit{<person> was awarded <false prize> for what specific reason?}'' and insert the corrupted triple ``\textit{<person>, was awarded, <false prize>}'' into the template.

Following this automatic construction pipeline, we construct two datasets, namely Prize and Movie. We choose the variants of Llama-2-Chat \citep{touvron2023llama} with 7B and 13B parameters as the triple selector. For each model within the dataset, different versions are constructed as varying numbers of knowledge triples are selected. Table \ref{dataset} provides the details of our datasets, while concrete question templates are presented in Appendix \ref{appendix:question_templates}. If the datasets were to be used with other models, researchers could readily follow our proposed pipeline and construct their own versions of the datasets tailored to their specific models.
\begin{table}[]
\resizebox{\linewidth}{!}{
\begin{tabular}{ccccc}
\rowcolor[HTML]{EDEDED} 
\hline
\multicolumn{2}{c}{\cellcolor[HTML]{EDEDED}\textbf{Dataset}} & \textbf{Knowledge}     & \cellcolor[HTML]{E7E6E6}\textbf{Selected} & \cellcolor[HTML]{E7E6E6}\textbf{Questions} \\
                                          & 7B                   &                        & 237                                                   & 948                                                  \\
\multirow{-2}{*}{Prize}                   & 13B                  & \multirow{-2}{*}{950}  & 457                                                   & 1828                                                 \\ \hline
                                          & 7B                   &                        & 1001                                                   & 4004                                                 \\
\multirow{-2}{*}{Movie}                   & 13B                  & \multirow{-2}{*}{5509} & 1001                                                  & 4004        \\ \hline                                        
\end{tabular}
}
\caption{Details of our datasets. The columns denote the dataset name, number of knowledge triples, number of selected knowledge triples for each model, number of constructed questions for each model, respectively. As we curate factual knowledge from specific models, two versions of each dataset are given.}
\label{dataset}
\end{table}


\section{Hallucination Analyze}
In this section, we conduct a comprehensive analysis of false premise hallucinations from the surface and delves deeper into the model step by step. 

\subsection{Analysis of Model Uncertainty}
\label{uncertainty}
In this part, we quantitatively investigate model uncertainty, which is a significant external feature of false premise hallucination and can be utilized to detect the hallucination occurrence. We hypothesize that model exhibits more inherent uncertainty when generating hallucinated answers. We design a model uncertainty measurement metric that allows the various linguistic forms of the true answer and experimental results validate our hypothesis.



\paragraph{Uncertainty Measurement} We utilize three metrics to measure the uncertainty of the model when confronted with a question. The former two out of the three metrics are straightforward while the third one is specifically designed for our task. Suppose that we have a question $q$ and a sequence of model answer $T=(t_0,t_1,...,t_N)$, the three metrics are described below:

(1) \textbf{PPL-Based} We simply calculate the negative log likelihood of the model answer: $U_1(q)=U_1(T)=-\frac{1}{|N|}\sum_{i=1}^{N}logp_{\theta}(x_i|x_{<i})$, where $logp_{\theta}(x_i|x_{<i})$ is the log likelihood of the $i$-th token based on the previous tokens $x_{<i}$.

(2) \textbf{Sampling-Based} To fully leverage the uncertainty in model parameters \citep{huang2023look-before-you-leap}, we generate multiple answer sequences $T_1,T_2,...,T_k$ for one question and calculate the average log likelihood across all sequences: $U_2(q)=\frac{1}{k}U_1(T_k)$.

(3) \textbf{Semantic-Based} Inspired by the incorporation of linguistic invariances in model uncertainty estimation \citep{semantic-uncertainty-iclr2023}, we separately treat the correct and incorrect answers among the multiple generated answer sequences. We consider all the correct answers as a unified semantic set and each incorrect answer as a discrete semantic set. Then we calculate the uncertainty over these semantic sets. Concretely, suppose that there are $K_1$ incorrect answers and $K_2$ correct answers in the $K$ generated answers, we calculate the uncertainty as follows: $U_3(q)=-\frac{1}{K_1+1}[\sum_{K_1}U_1(T_k)+ log \sum_{K_2}expU_1(T_k)]$.


\begin{figure}[t]
\centering
\subfloat[7B]{\includegraphics[width=0.5\linewidth]{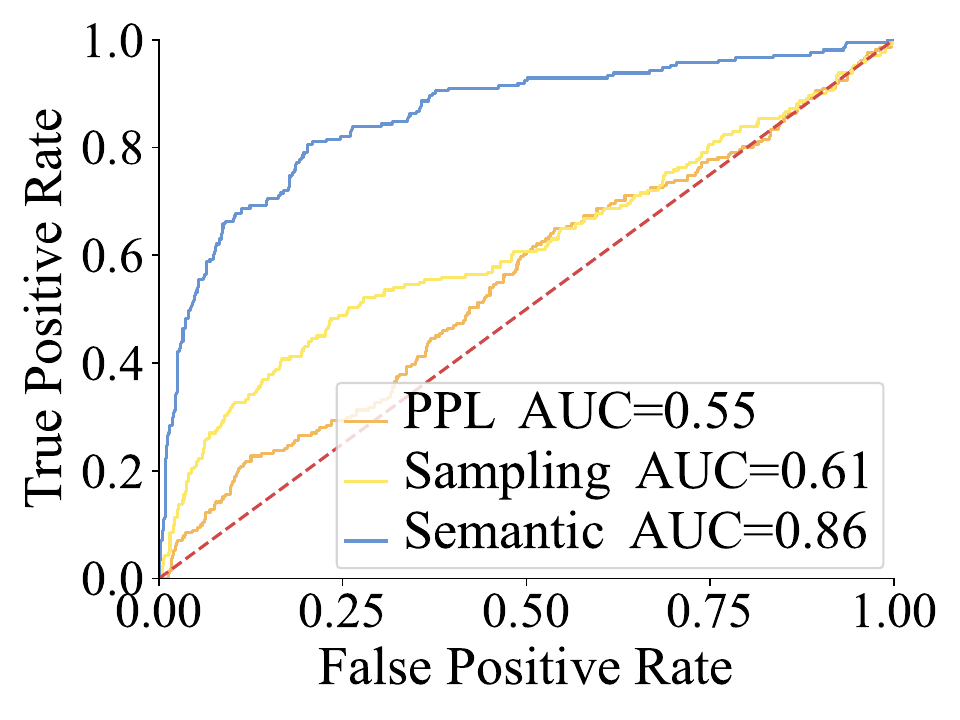}\label{attn_head_influence}} \hfill
\subfloat[13B]{\includegraphics[width=0.5\linewidth]{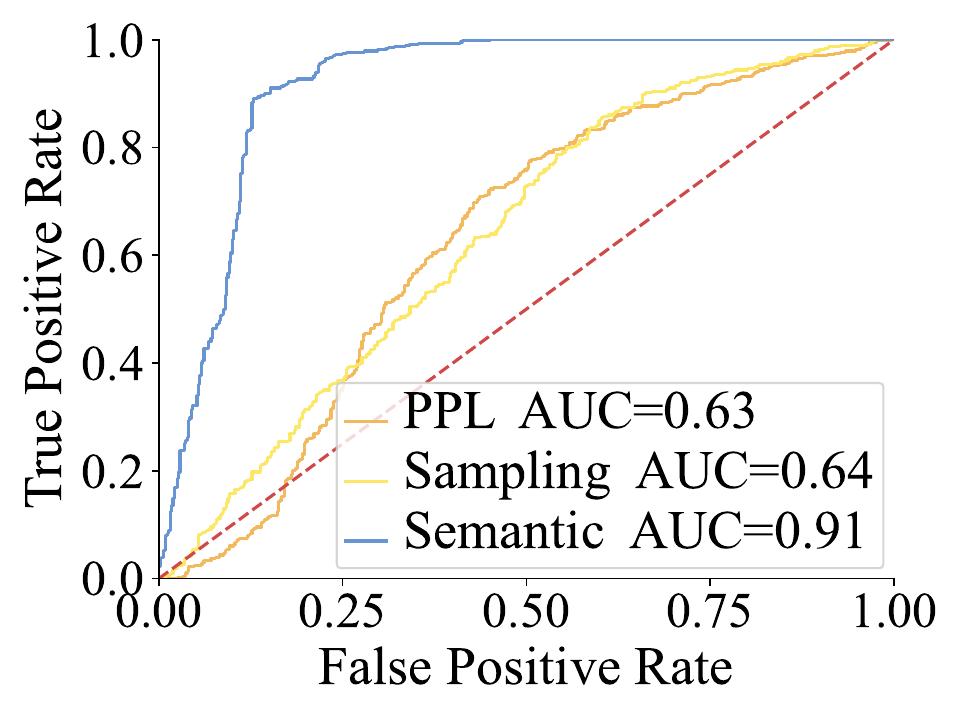}\label{attn_head_influence}} 
\caption{The Receiver Operating Characteristic Curve on the Movie dataset. The perfect AUC score is 1 while the random AUC score is 0.5.}
\label{uncertainty_result}
\end{figure}

\paragraph{Experiment} We conduct experiments on the Movie dataset using Llama-2-7b-chat (denoted as 7B) and Llama-2-13b-chat (denoted as 13B). To evaluate the model uncertainty during hallucination occurrences, the calculated uncertainty scores are used for the binary classification task, aimed at determining the occurrence of hallucinations for each false premise question. We use the Area Under the Receiver Operating Characteristic (AUC) to assess the effectiveness of the uncertainty score. The higher the scores, the greater the correlation between the uncertainty metric and the occurrence of hallucinations.

\paragraph{Results and Analysis}
The Receiver Operating Characteristic Curve (ROC Curve) curves and the AUC scores are shown in Figure \ref{uncertainty_result}. From the results, we draw the following observations: (1) Our semantic-based uncertainty metric score is far more effective than the other two methods. It can be further employed in the prediction of the occurrence of false premise hallucinations without relying on an external knowledge base. (2) We observe a strong correlation (over $0.9$ for the 13B model) between the occurrence of hallucinations and model uncertainty. This verifies our hypothesis that models exhibit more inherent uncertainty when generating hallucinated answers.




\begin{figure}[t]
\centering
\subfloat[7B w/ Hallucination]{\includegraphics[width=0.5\linewidth]{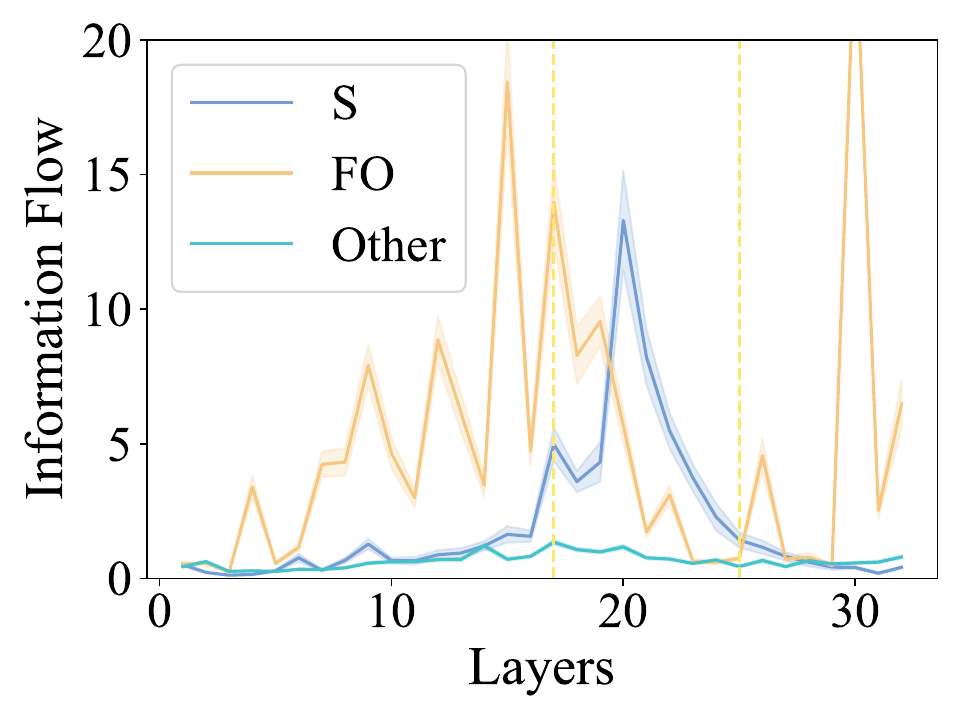}\label{7b_w}} \hfill
\subfloat[7B w/o Hallucination]{\includegraphics[width=0.5\linewidth]{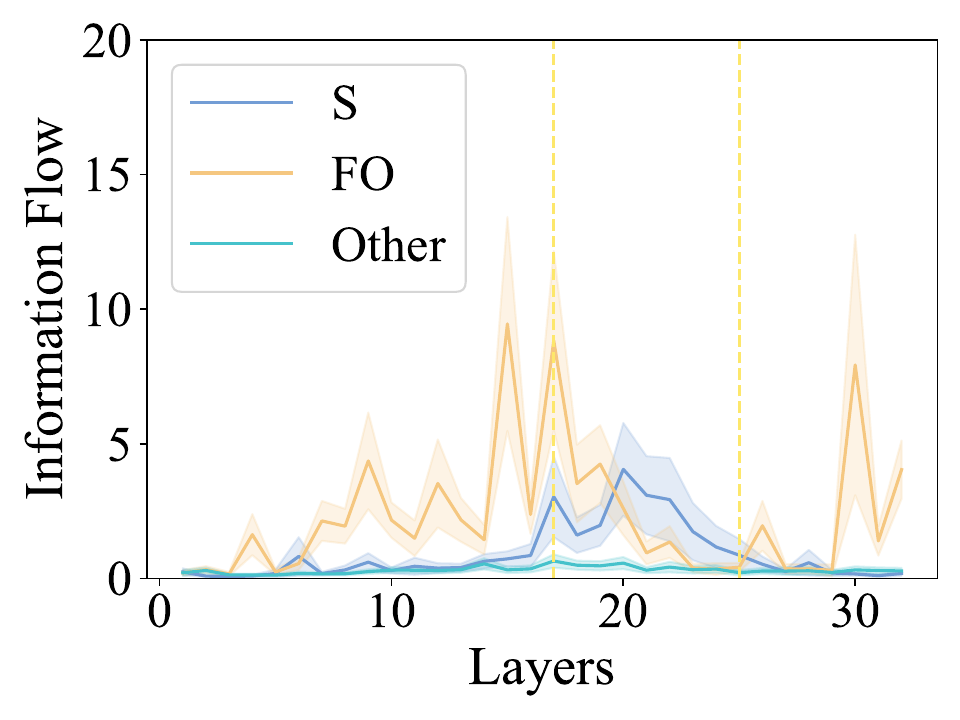}\label{7b_wo}} \hfill
\subfloat[13B w/ Hallucination]{\includegraphics[width=0.5\linewidth]{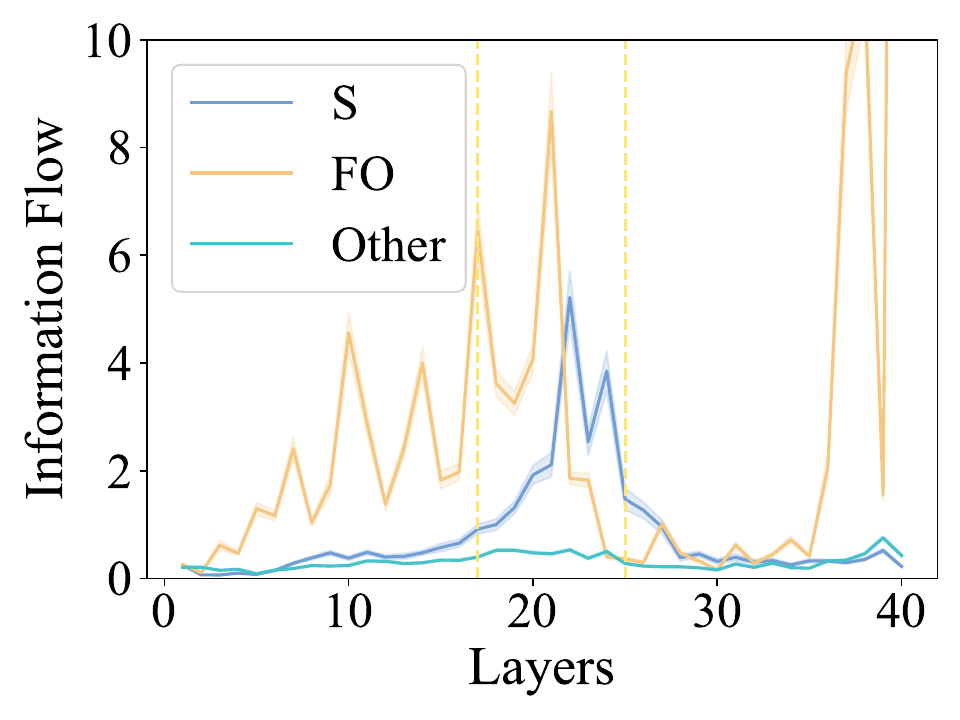}\label{13b_w}} \hfill
\subfloat[13B w/o Hallucination]{\includegraphics[width=0.5\linewidth]{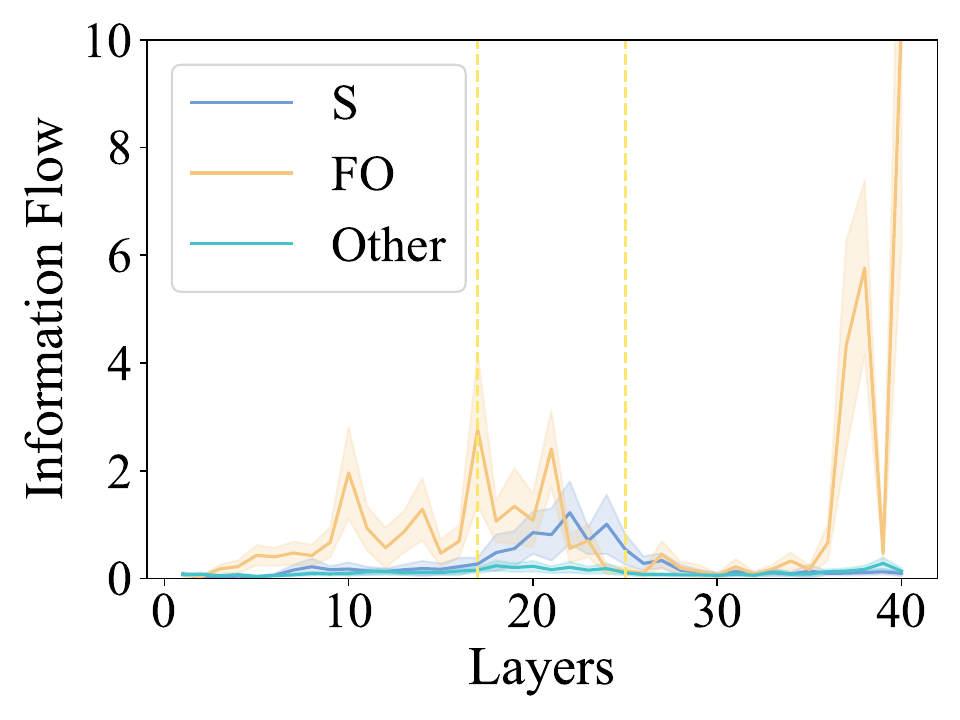}\label{13b_wo}} 
\caption{Information flow from various parts of the question to the final logit across distinct layers on hallucinated and non-hallucinated samples. }
\label{info_flow_result}
\end{figure}

\subsection{Analysis of Internal Information Flow}
\label{info_flow_section}

To explore the source of the uncertainty, in this section, we delve into the internal information flow of LLMs when generating hallucinated answers. We study how the information flow from different parts of the false premise question in the fill-in-the-blank task. Experimental results demonstrate that the knowledge about the subject is disturbed in the shallow layers of the model, particularly around the false object mentioned in the question.


\paragraph{Knowledge Assessment Task} 
The most significant feature of false premise hallucination is that the factual knowledge about the subject can be recalled directly. Therefore, we design a fill-in-the-blank task to evaluate how the knowledge stored in the model parameters is affected by the question. Concretely, for a question containing the false triple $(s,r,o')$, LLMs are required to complete the following cloze query: ``\textit{<Question> According to my knowledge, the object linking from subject $s$ via relation $r$ is \textunderscore}''. We posit that this knowledge assessment task is correlated with the original question answering task. Intuitively, if LLMs generate hallucinated answers to false premise questions, they are highly likely to fail in completing this cloze query correctly.


\paragraph{Attribution Score} We aim to discover how the information flow from the tokens in the question to the final prediction logit in the knowledge assessment task. Since gradients and the attention itself can be blended together to acquire a better performance \citep{ExplainabilityForLLMSurvey}, we use the element-wise product version \citep{wang-etal-2023-label-words-are-anchors} to calculate the attribution score for each token:
\begin{equation*}
    S_l = \left| \sum_{h=1}^{H} A^h_l \odot \frac{\partial L}{\partial A^h_l} \right|
\end{equation*}
where $A_l^h$ is the attention pattern matrix described in Section \ref{transformer_architecture} and $L$ is the loss on the token prediction task. The attribution score matrix $S_l$ on layer $l$ is a $N\times N$ matrix ($N$ is the length of the prompt). We partition the question into three parts: subject part (denote as $S$), false object part (denote as $FO$) and other part (denoted as $other$). The information flow from these parts is consequently defined as:
\begin{equation*}
\begin{aligned}
S_{S_l} &= \frac{1}{\lvert N_{sub} \rvert} \sum_{t=1}^{N_{sub}} S_{N,i_t} \\
S_{FO_l} &= \frac{1}{\lvert N_{fobj} \rvert} \sum_{t=1}^{N_{fobj}} S_{N,i_t} \\
S_{other_l} &= \frac{1}{\lvert N_{other} \rvert} \sum_{t=1}^{N_{other}} S_{N,i_t}
\end{aligned}
\end{equation*}
where $N_{sub}$ is the number of tokens of subject, $N_{fobj}$ is the number of tokens of the false object and $N_{other}$ is the number of other tokens.


\paragraph{Results and Analysis.} We illustrate the information flow from various parts of the question to the final logit across distinct layers on hallucinated and non-hallucinated samples in the Prize dataset, as shown in Figure \ref{info_flow_result}. By observing the figures, it's evident that the information flow across the layers can be roughly divided into three pieces. 
(1) In shallow layers, models primarily focus on the false object part of the question, leading to more perturbation on the hallucinated samples than the non-hallucinated samples.
(2) In middle layers, to counteract the perturbation caused by the false object part, models shift their emphasis towards the subject to validate the knowledge. The resistance observed in the hallucinated examples is greater than in the non-hallucinated samples. 
(3) In deep layers, the models continue to focus on the false object component of the question. 

Therefore, we conclude that the knowledge about the subject is disturbed in the shallow layers of the model in the false object part of the question.



\begin{figure}
\begin{center}
\resizebox{\linewidth}{!}{
\includegraphics[]{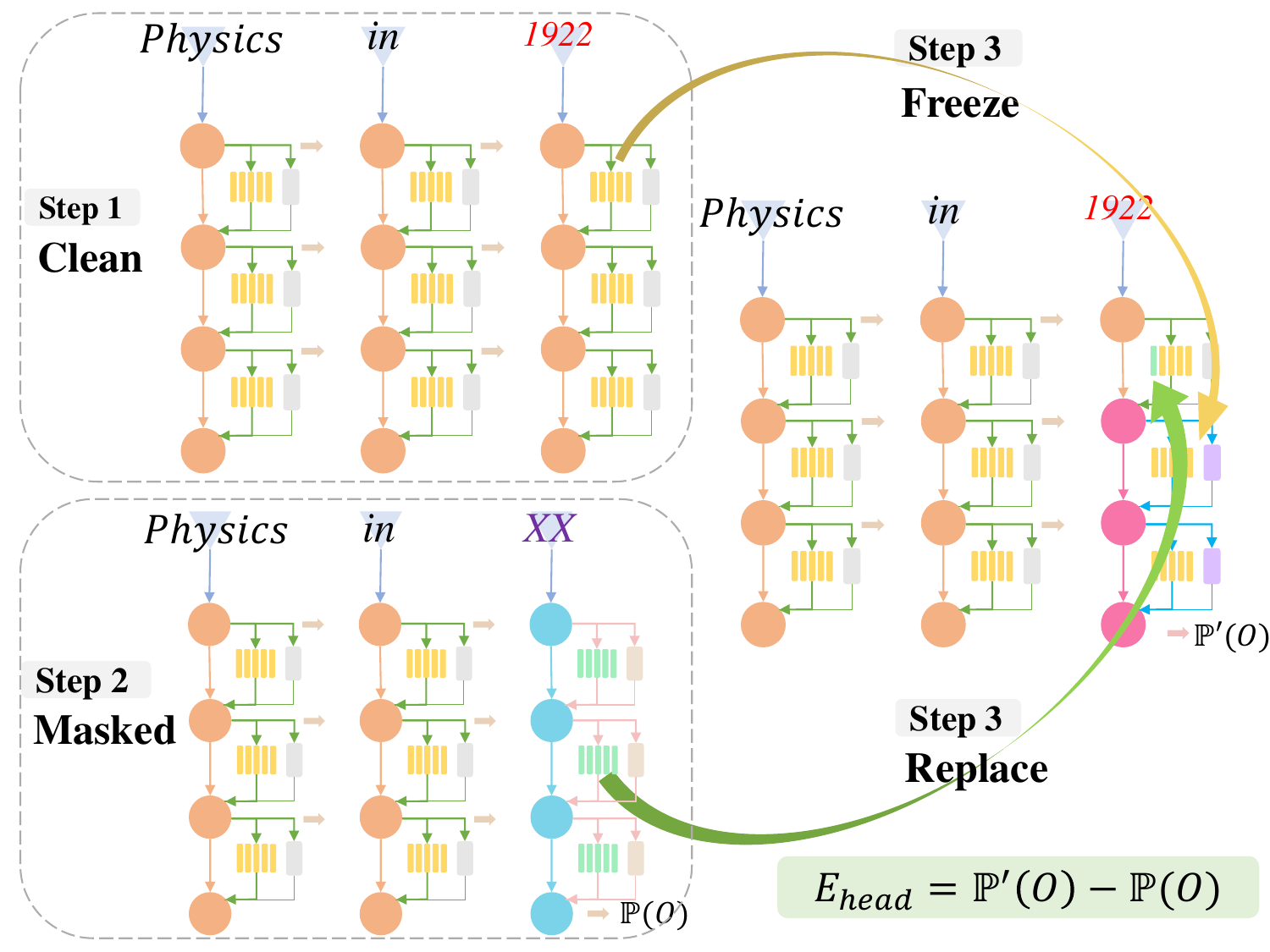} 
}
\caption{Calculation of the influence of a single attention head.}
\label{path_patch}
\end{center}
\end{figure}

\begin{figure*}
\centering
\subfloat[Attention Head Influence]{\includegraphics[width=0.66\columnwidth]{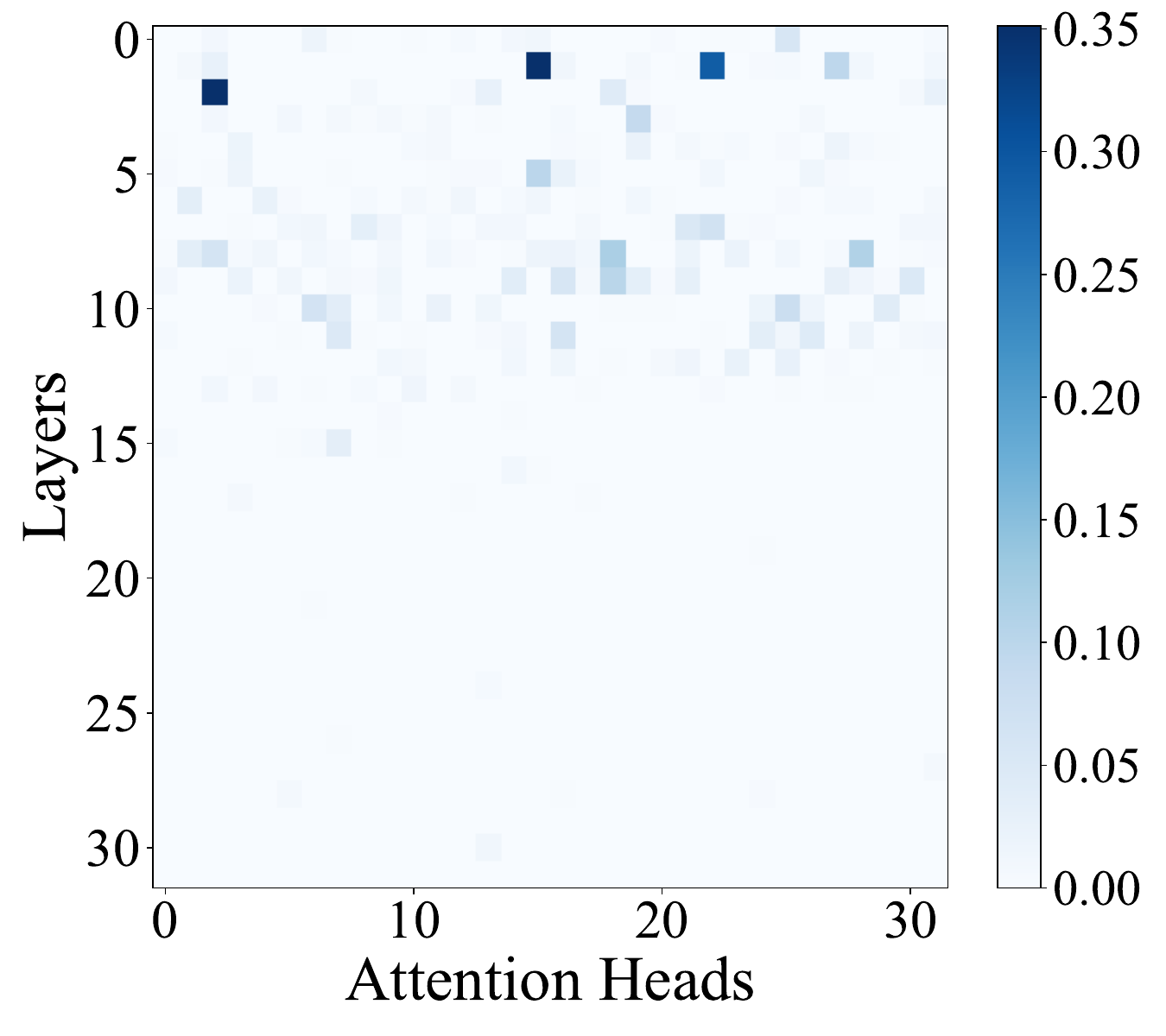}\label{attn_head_influence}}\hspace{.2in}
\subfloat[Attention Pattern of head 1-22.]{\includegraphics[width=0.66\columnwidth]{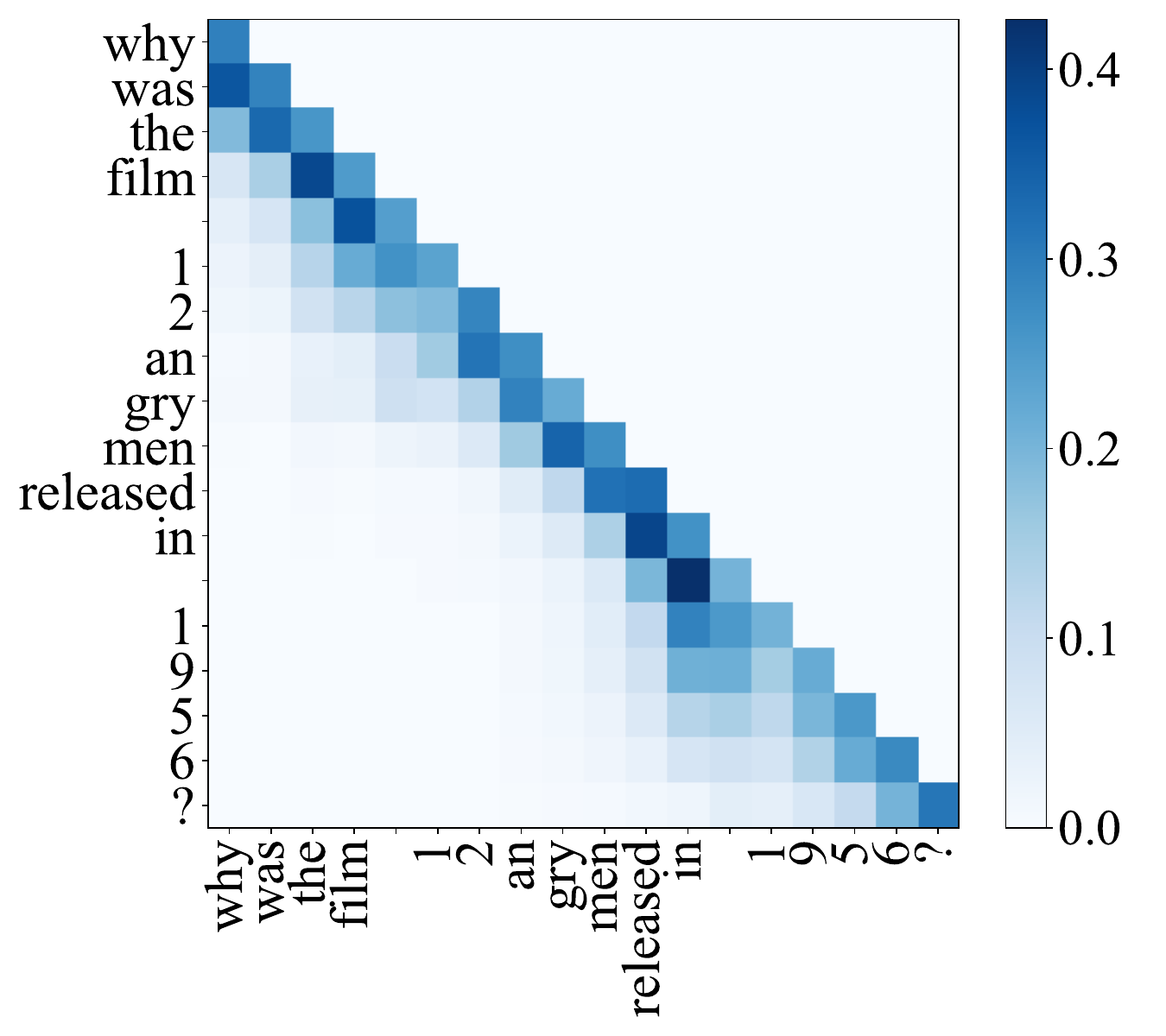}\label{head_1_22}}
\subfloat[Attention Pattern of head 5-15.]{\includegraphics[width=0.66\columnwidth]{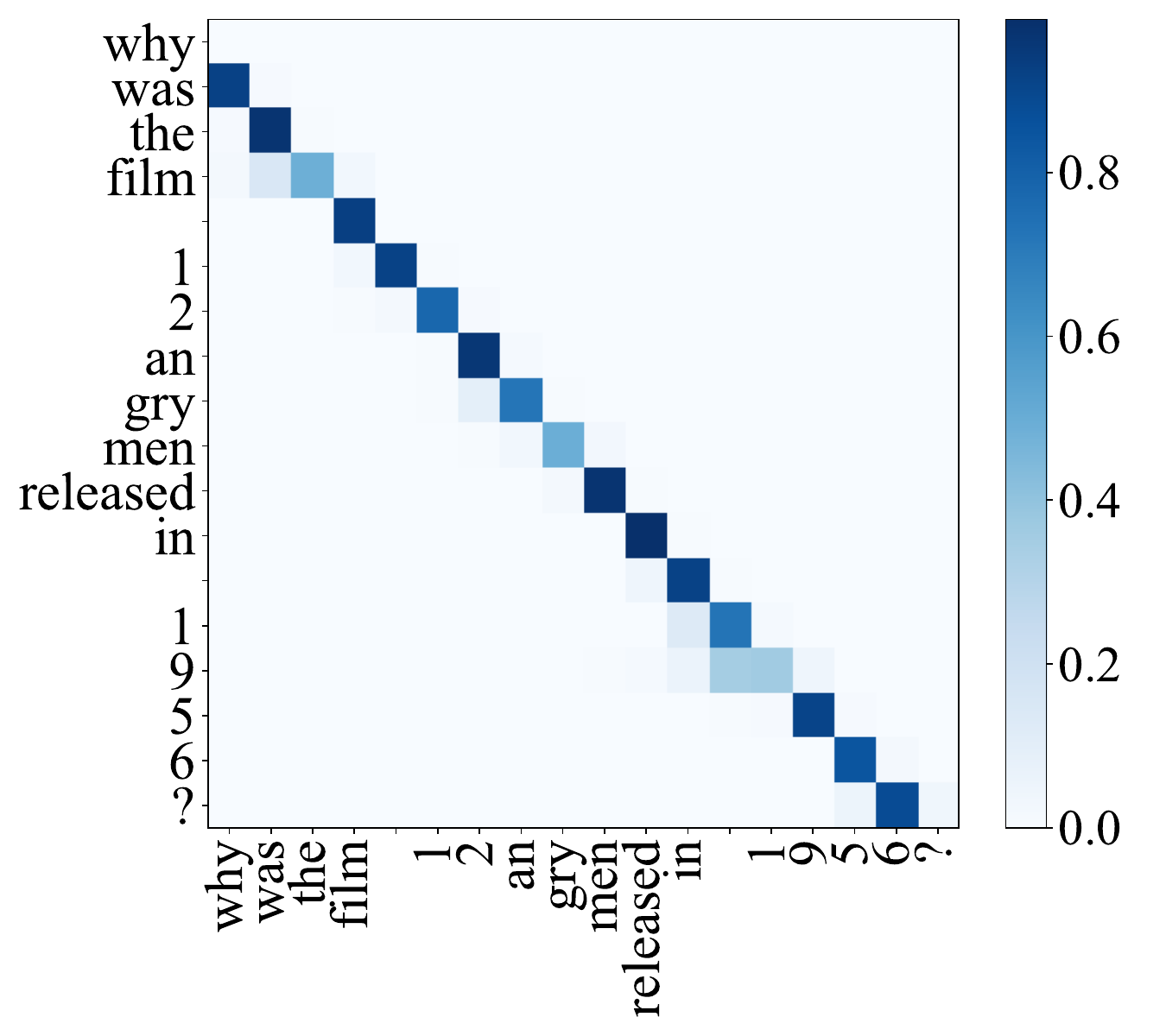}\label{head_5_15}}

\caption{Illustration of the false premise attention heads. }
\label{path_patch_result}
\end{figure*}

\subsection{Analysis of Individual Attention Heads}
\label{find_fp_heads}
As many studies \citep{meng2022locating-and-editing,yuksekgonul2023attention-satisfies} indicate that the self-attention layers transfer the factual knowledge stored in the MLP layers during the inference process, we further investigate the influence of each individual attention head within the self-attention layers to identify the source of the disturbance. We calculate the influence of each individual head on the final prediction logit in the knowledge assessment task and find out the presence of false premise attention heads.


\paragraph{Influence Calculation} We propose a method to investigate the influence of an individual attention head on the prediction logit in the knowledge assessment task. The computation of the influence of a specific individual attention head can be divided into three steps, as shown in Figure \ref{path_patch}.

(1) \textbf{Clean Run.} We perform a forward pass using the original question and store the activations of all the attention heads. The token prediction logit of this run is denoted as $\mathbb{P(O)}$.

(2) \textbf{Masked Run.} We create a masked question by substituting the false object tokens with nonsensical placeholders. As illustrated in the top-left of Figure \ref{path_patch}, the false year is substituted with `\textit{XX}'.  Subsequently, we perform a forward pass using the masked question to store the activations of all the attention heads. 

(3) \textbf{Replace and Freeze Run.} We run another forward pass using the original question,  replacing the selected attention head with the values stored in the masked run while simultaneously freezing other attention heads using the values stored in the clean run. The token prediction logit is denoted as $\mathbb{P'(O)}$.
Therefore, the influence of a specific attention head can be defined as $E_{head}=\mathbb{P'(O)}-\mathbb{P(O)}$.


\paragraph{Results and Analysis} 
We visualize the influence of the attention heads in Llama-2-7b-chat averaged across all samples in the Movie dataset, as depicted in Figure \ref{attn_head_influence}. From the figure, we draw the following key observations: (1) The attention heads that exert great influence on the final logit primarily reside in the shallow layers of the model (0-15 layers for the 7B model). This is consistent with the analysis of the internal information flow which indicates that the disturbance of factual knowledge originates from the shallow layers, as discussed in Section \ref{info_flow_section}. (2) We observe that a few attention heads exert significantly greater influence than others. We conclude that these attention heads will have to take the blame for the false premise hallucination and we designate them as \textbf{False Premise Attention Heads}. 

\paragraph{Attention Patterns} In order to better understand the behaviour of the false premise attention heads, we explore their attention patterns in some concrete examples. Figure \ref{head_1_22} and \ref{head_5_15} shows the attention pattern of the 23-rd attention head in the second layer (denoted as (1,22)) and the 16th head in the sixth layer (denoted as (5,15)) on two examples in the Movie dataset. It is evident that they both exhibit a similar pattern, primarily concentrating on the information around the current tokens while disregarding the connection with other tokens.

Therefore, the internal working mechanism of false premise hallucination is revealed: the false premise heads solely focus on the information surrounding the current tokens, disregarding the connection between the false object and the subject, which contributes to the occurrence of the false premise hallucination.

\section{Hallucination Mitigation}
\label{hallucination_mitigation}
Based on our previous in-depth analysis, in this section, we introduce FAITH, a novel method aimed at mitigating false premise hallucinations.

\IncMargin{1em} 
\begin{algorithm}[t!] \SetKwData{Left}{left}\SetKwData{This}{this}\SetKwData{Up}{up} \SetKwFunction{Union}{Union}\SetKwFunction{FindCompress}{FindCompress} \SetKwInOut{Input}{input}\SetKwInOut{Output}{output}
	
	\Input{A set of false premise questions $\{Q_1,Q_2,...,Q_N\}$, model $M$, threshold $\tau$, question templates $\{T_1,T_2,...,T_N\}$.}
	\Output{A set of attention heads in $M$}
	 \BlankLine

	 $S \longleftarrow$ []\;
      $L \longleftarrow$ NumberLayersOf($M$)\;
      $H \longleftarrow$ NumberHeadsOf($M$)\;
	 \For{$i\leftarrow 1$ \KwTo $N$}{ 
            $p_i \leftarrow (Q_i,T_i)$  \;
            \For{$l\leftarrow 1$ \KwTo $L$}{
                \For{$h\leftarrow 1$ \KwTo $H$}{
                    $E_{lh} \leftarrow$ Influence($l,h,p_i$)\;
                    \If{$E_{lh} \ge \tau$}{
                        $S \leftarrow (l,h)$\;
                    }
                }
            }
 	}
      $S \longleftarrow$ SortedByFrequency($S$)\;
 	 	  \caption{Head Localization}
 	 	  \label{alg_localization} 
 	 \end{algorithm}
 \DecMargin{1em}

\begin{table*}[]
\resizebox{\linewidth}{!}{
\begin{tabular}{cc|ccccc|ccccc}
\hline
 \arrayrulecolor{gray!100}
\multicolumn{2}{c|}{\multirow{2}{*}{Methods}} & \multicolumn{5}{c|}{Prize} & \multicolumn{5}{c}{Movie} \\
\multicolumn{2}{c|}{}                         & T1  & T2  & T3 & T4 & Avg & T1  & T2  & T3 & T4 & Avg \\ \hline
\multirow{5}{*}{7B}       & Vanilla                  & 14.77 & 15.19 & 13.08 & 14.35 & 14.35 & 78.92 & 41.46 & 41.66 & 24.08 & 46.53 \\
& ITI                      & 36.71 & 14.77 & \textbf{18.99} & 22.36 & 23.21 & 74.63 & 43.16 & 30.87 & 10.69 & 39.84 \\
& DoLA                     & 17.30 & 19.41 & \textbf{18.99} & 27.00 & 20.68 & 72.53 & 59.24 & 38.86 & 35.46 & 51.52 \\
& Repe                     & 23.21 & 25.74 & 8.86  & 25.74 & 20.89 & 47.15 & 49.15 & 40.56 & 42.06 & 44.73 \\  \rowcolor{gray!7}
& FAITH (Ours)              & \textbf{62.03} & \textbf{38.40 }& 18.14 & \textbf{35.86} & \textbf{38.61} & \textbf{94.31} & \textbf{81.02} & \textbf{65.10} & \textbf{70.53} & \textbf{77.74}  \\ \hline
\multirow{5}{*}{13B}      & Vanilla                  & 8.32  & 4.38  & 2.41 & 3.94  & 4.76    & 52.05 & 42.06 & 2.30  & 0.80  & 24.30   \\
& ITI                      & 9.19  & 6.78  & 2.41 & 4.81  & 5.80    & 74.93 & 67.83 & 4.30  & 2.80  & 37.47   \\
& DoLA                     & 21.44 & 18.16 & 1.97 & 4.60  & 11.54   & 56.84 & 58.74 & 2.40  & 2.80  & 30.20   \\
& Repe                     & 19.47 & \textbf{39.17} & 7.44 & 28.45 & 23.63   & 51.35 & 44.16 & 17.48 & 21.58 & 33.64   \\ \rowcolor{gray!7}
& FAITH (Ours)                     & \textbf{30.63} & 18.16 & \textbf{8.97} & \textbf{39.61}& \textbf{24.34}   & \textbf{82.02} & \textbf{80.52 }& \textbf{49.65 }& \textbf{23.78} & \textbf{58.99} \\  \arrayrulecolor{black}  \hline
\end{tabular}
}
\caption{Experimental results (accuracy, \%) using Llama-2-7b-chat and Llama-2-13b-chat. The best results are highlighted in \textbf{boldface}. ``T1'' denotes the model performance on question template 1. ``Avg" denotes the model performance averaged across all four question templates. A higher presented accuracy indicates a lower occurrence of hallucination.} 
\label{7b_and_13b_result}
\end{table*}

\subsection{FAITH} 
Our method FAITH consists of two parts. The first part involves localizing the false premise attention heads of a set of false premise questions, while the second part involves constraining these attention heads during the model inference process.

\paragraph{Head Localization} To identify the false premise heads for a set of false premise questions, for each question, we employ the knowledge assessment cloze queries to convert the question answer task into the fill-in-the blank task.  Subsequently, we calculate the influence of each individual attention head on each specific sample, as described in Section \ref{find_fp_heads}. Finally, we select attention heads that have an influence exceeding a predefined threshold on individual examples and appear most frequently across the samples. The pseudocode of our localization procedure is shown in Algorithm \ref{alg_localization}.

\paragraph{Head Constraining} To eliminate the impact of these false premise heads and mitigate hallucinations, we constrain these attention heads around the false object tokens during the model inference process. 
The output of the constrained multi-head attention is defined as:
\begin{equation*}
\begin{aligned}
         a_{l}' &= \sum_{h=1}^{H} A^h (x_{l-1} W^{h}_V) W^{h}_{O} \odot f(\mathbf{1}_{N \times d}) \\
         f(B) &= 
         \begin{cases}
            B[i:j, :]=0, & \text{if } h \in S, \\
            B, & \text{else}.
         \end{cases}
\end{aligned}
\end{equation*}
 
where $S$ is the set of false premise heads on layer $l$, where $W_K^h,W^h_Q,W^{h}_{O} \in \mathbb{R}^{d\times d_h},W^{h}_{O} \in \mathbb{R}^{d_h\times d}$ are the parameter matrices, $A \in \mathbb{R}^{N\times N}$ is the attention pattern matrix, $a_l'$ is the output of the constrained multi-head attention, $i,j$ is the range of the false object tokens in the question and $f(B)$ is the constraining function, which zeroes out certain rows of the input matrix if the attention head is to be constrained. The rows of the matrix $\mathbf{1}_{N \times d}$ correspond to the tokens in the question thus we choose the false object part of the question to eliminate.



\subsection{Baselines}
We compare our method with the following baseline methods:

(1) \textbf{Vanilla}, which directly prompts the LLMs to generate the answers without any intervention.

(2) \textbf{ITI} \citep{li2023inference-time-intervention}, which is a technique that adjusts certain attention heads towards the `truthful' direction during the inference process.

(3) \textbf{DoLa} \citep{chuang2023dola}, which is a novel decoding strategy that better reveals the truthful knowledge by contrasting different layers.

(4) \textbf{RepE} \citep{zou2023transparency}, which computes the difference vector using a pair of contrastive prompts during inference and utilizes it to control the hidden state during the inference process. 

\subsection{Implementation Details}
We conduct experiments using open source LLMs, specifically Llama-2-7b-chat and Llama-2-13b-chat, on both the Movie and Prize dataset. To prevent errors in a single decoding step, we employ beam search decoding and set the beam size to $5$. For 7B model on both the datasets, we constrain 5 false premise attention heads (approximately $0.56\%$ of all the attention heads). For 13B model, we constrain 15 ($0.94\%$) false premise attention heads on the Movie dataset and 20 ($1.25\%$)  on the Prize dataset.

For the evaluation metrics of the hallucination mitigation task, we employ a heuristic method. We consider the answer to a false premise question as non-hallucinated if the original object $o$ is present in the final answer. This indicates that LLMs have successfully identified the false premise in the question. Therefore, accuracy can be employed as the metric to measure the performance of each method. The higher the accuracy, the lower the occurrence of hallucination.

\subsection{Results and Analysis}

From the experimental results shown in Table \ref{7b_and_13b_result}, we derive the following key observations. (1) Our method is considerably effective when compared with existing baselines. For example, our method consistently outperforms other baselines across nearly all the question templates. This verifies the hypothesis that false premise heads contribute to model hallucinations. (2) Our method is more effective on models with smaller number of parameters. For example, compared with the second best-performing method on the Prize dataset, our method achieves $17.72\%$ improvements of accuracy with the 7B model yet $0.71\%$ improvements with the 13B model. We attribute it to that models with larger scales tend to exhibit greater resistance to changes in their results.




\subsection{Generalization}
We further explore the generalizability of the identified false premise attention heads from one question template to others. We design the following two experiments: (1) \textbf{Within Knowledge}, which uses the false premise heads identified on various question templates in Movie dataset to mitigate hallucinations on each specific template. (2) \textbf{Across Knowledge} ,which uses the false premise heads identified on the Prize dataset to mitigate hallucinations on the Movie dataset. 



We also choose random selected attention heads for comparison and the experimental results are shown in Table \ref{generalization_7b}. From the table, we can observe that our identified false premise attention heads exhibit strong generalizabilities. For example, the model achieves comparable performance within and across datasets and significantly outperforms the random baseline. This demonstrates that our revealed mechanism of the false premise attention heads is relatively general.




\begin{table}[]
\resizebox{\linewidth}{!}{
\begin{tabular}{cr|ccccc}
\hline
    \arrayrulecolor{gray!100}
\multicolumn{2}{c|}{Methods}                              & T1    & T2    & T3    & T4    & Avg   \\ \hline
\multicolumn{2}{c|}{FAITH}                                & 94.31 & 81.02 & 65.10 & 70.53 & \textbf{77.74} \\
\rowcolor[HTML]{E7E6E6} 
\cellcolor[HTML]{E7E6E6}                         & w/T1  & 94.31 & 83.02 & 61.54 & 68.73 & 76.90 \\
\rowcolor[HTML]{E7E6E6} 
\cellcolor[HTML]{E7E6E6}                         & w/T2  & 94.51 & 81.02 & 66.83 & 70.93 & \textbf{78.32} \\
\rowcolor[HTML]{E7E6E6} 
\cellcolor[HTML]{E7E6E6}                         & w/T3  & 88.91 & 61.74 & 65.1  & 43.26 & 64.75 \\
\rowcolor[HTML]{E7E6E6} 
\multirow{-4}{*}{\cellcolor[HTML]{E7E6E6}Within} & w/T4  & 94.31 & 81.02 & 60.64 & 70.53 & 76.63 \\
                                                 & w/PT1 & 94.11 & 80.22 & 65.53 & 69.53 & \textbf{77.35} \\
                                                 & w/PT2 & 94.11 & 80.22 & 65.53 & 69.53 & \textbf{77.35 }\\
                                                 & w/PT3 & 94.71 & 81.82 & 63.64 & 68.63 & 77.20 \\
\multirow{-4}{*}{Across}                         & w/PT4 & 94.31 & 81.02 & 60.64 & 70.53 & 76.63 \\
\rowcolor[HTML]{E7E6E6} 
\multicolumn{2}{c|}{\cellcolor[HTML]{E7E6E6}Random}       & 78.32 & 41.46 & 41.06 & 23.88 & \textbf{46.18 }  \\  \arrayrulecolor{black} \hline
\end{tabular}
}
\caption{Generalizability of the attention heads on the 7B model. ``w/T1'' denotes using the false premise heads identified on the question template 1 in the same Movie dataset. ``w/PT1"" denotes using the false premise heads identified on the question template 1 in the Prize dataset. Results of the 13B model can be found in Appendix \ref{appendix:generalization}.}
\label{generalization_7b}
\end{table}







\section{Related Work}

\paragraph{Hallucination}
Many work focus on evaluating \citep{vu2023freshllms,li2023halueval}, detecting \citep{chen2024inside,yang-etal-2023-new-benchmark-phd} and mitigating \citep{trivedi-etal-2023-interleaving-retrieval,gao-etal-2023-rarr,mündler2023selfcontradictory,zhou2023analyzing} hallucinations. However, they ignore the analysis of the false premise hallucination.

\paragraph{Mechanistic Interpretability} Mechanistic interpretability aims at understanding the model behaviours by investigating individual neurons and their connections \citep{ExplainabilityForLLMSurvey}. Various interpretable representations are found, such as in model alignment \cite{lee2024mechanistic_alignment}, reasoning \citep{stolfo2023mechanistic_reasoning}, knowledge recall \citep{geva-etal-2023-dissecting,yu-etal-2023-characterizing} and in-context-learning\citep{hendel-etal-2023-context,todd2023function_vectors}. We are the first to explore the internal working mechanism of false premise hallucinations.

\section{Conclusion}
In this paper, we conduct a comprehensive analysis of an important type of hallucination: False Premise Hallucination. Our analysis begins at the surface of the model and gradually delves deeper into it, ultimately revealing the presence of false premise attention heads. Based on our analysis, we propose a novel false premise hallucination mitigation method, \textbf{FAITH} (\textbf{F}alse premise \textbf{A}ttention head constra\textbf{I}ning for mi\textbf{T}igating \textbf{H}allucinations). Extensive experiments demonstrate the effectiveness of our method comparing with the baselines and the promising nature of our revealed internal working mechanism of false premise hallucination.
\section*{Limitations}
Our study, while providing valuable insights into the false premise hallucination, is subject to several limitations, as outlined below. (1) Due to constraints in computing resources, our research is restricted to models up to a scale of 13B parameters. Future research could investigate more models with larger scales. (2) The calculation of the influence of multiple attention heads is time-consuming due to the vast number of combinations. Consequently, considering the computational complexity involved, we limit our investigation to the influence of each individual attention head on the final prediction logit. Future research could further explore how to effectively select the the most influential joint contribution of multiple attention heads.




\bibliography{custom}

\begin{thebibliography}{41}
\expandafter\ifx\csname natexlab\endcsname\relax\def\natexlab#1{#1}\fi

\bibitem[{Bang et~al.(2023)Bang, Cahyawijaya, Lee, Dai, Su, Wilie, Lovenia, Ji, Yu, Chung, Do, Xu, and Fung}]{bang2023multitask-chatgpt-evaluation}
Yejin Bang, Samuel Cahyawijaya, Nayeon Lee, Wenliang Dai, Dan Su, Bryan Wilie, Holy Lovenia, Ziwei Ji, Tiezheng Yu, Willy Chung, Quyet~V. Do, Yan Xu, and Pascale Fung. 2023.
\newblock \href {https://aclanthology.org/2023.ijcnlp-main.45} {A multitask, multilingual, multimodal evaluation of {C}hat{GPT} on reasoning, hallucination, and interactivity}.
\newblock In \emph{Proceedings of the 13th International Joint Conference on Natural Language Processing and the 3rd Conference of the Asia-Pacific Chapter of the Association for Computational Linguistics (Volume 1: Long Papers)}, pages 675--718, Nusa Dua, Bali. Association for Computational Linguistics.

\bibitem[{Bubeck et~al.(2023)Bubeck, Chandrasekaran, Eldan, Gehrke, Horvitz, Kamar, Lee, Lee, Li, Lundberg, Nori, Palangi, Ribeiro, and Zhang}]{bubeck2023sparks_of_agi}
Sébastien Bubeck, Varun Chandrasekaran, Ronen Eldan, Johannes Gehrke, Eric Horvitz, Ece Kamar, Peter Lee, Yin~Tat Lee, Yuanzhi Li, Scott Lundberg, Harsha Nori, Hamid Palangi, Marco~Tulio Ribeiro, and Yi~Zhang. 2023.
\newblock \href {http://arxiv.org/abs/2303.12712} {Sparks of artificial general intelligence: Early experiments with gpt-4}.

\bibitem[{Carlini et~al.(2023)Carlini, Ippolito, Jagielski, Lee, Tram{\`{e}}r, and Zhang}]{carlini2022quantifying-memorization}
Nicholas Carlini, Daphne Ippolito, Matthew Jagielski, Katherine Lee, Florian Tram{\`{e}}r, and Chiyuan Zhang. 2023.
\newblock \href {https://openreview.net/pdf?id=TatRHT\_1cK} {Quantifying memorization across neural language models}.
\newblock In \emph{The Eleventh International Conference on Learning Representations, {ICLR} 2023, Kigali, Rwanda, May 1-5, 2023}. OpenReview.net.

\bibitem[{Chen et~al.(2024)Chen, Liu, Chen, Gu, Wu, Tao, Fu, and Ye}]{chen2024inside}
Chao Chen, Kai Liu, Ze~Chen, Yi~Gu, Yue Wu, Mingyuan Tao, Zhihang Fu, and Jieping Ye. 2024.
\newblock \href {http://arxiv.org/abs/2402.03744} {Inside: Llms' internal states retain the power of hallucination detection}.

\bibitem[{Chuang et~al.(2023)Chuang, Xie, Luo, Kim, Glass, and He}]{chuang2023dola}
Yung-Sung Chuang, Yujia Xie, Hongyin Luo, Yoon Kim, James Glass, and Pengcheng He. 2023.
\newblock Dola: Decoding by contrasting layers improves factuality in large language models.
\newblock \emph{arXiv preprint arXiv:2309.03883}.

\bibitem[{Elhage et~al.(2021)Elhage, Nanda, Olsson, Henighan, Joseph, Mann, Askell, Bai, Chen, Conerly et~al.}]{elhage2021mathematical-framework-for-transformer-circuits}
Nelson Elhage, Neel Nanda, Catherine Olsson, Tom Henighan, Nicholas Joseph, Ben Mann, Amanda Askell, Yuntao Bai, Anna Chen, Tom Conerly, et~al. 2021.
\newblock A mathematical framework for transformer circuits.
\newblock \emph{Transformer Circuits Thread}, 1.

\bibitem[{Fan et~al.(2019)Fan, Jernite, Perez, Grangier, Weston, and Auli}]{fan2019eli5}
Angela Fan, Yacine Jernite, Ethan Perez, David Grangier, Jason Weston, and Michael Auli. 2019.
\newblock \href {https://doi.org/10.18653/v1/P19-1346} {{ELI}5: Long form question answering}.
\newblock In \emph{Proceedings of the 57th Annual Meeting of the Association for Computational Linguistics}, pages 3558--3567, Florence, Italy. Association for Computational Linguistics.

\bibitem[{Gao et~al.(2023)Gao, Dai, Pasupat, Chen, Chaganty, Fan, Zhao, Lao, Lee, Juan, and Guu}]{gao-etal-2023-rarr}
Luyu Gao, Zhuyun Dai, Panupong Pasupat, Anthony Chen, Arun~Tejasvi Chaganty, Yicheng Fan, Vincent Zhao, Ni~Lao, Hongrae Lee, Da-Cheng Juan, and Kelvin Guu. 2023.
\newblock \href {https://doi.org/10.18653/v1/2023.acl-long.910} {{RARR}: Researching and revising what language models say, using language models}.
\newblock In \emph{Proceedings of the 61st Annual Meeting of the Association for Computational Linguistics (Volume 1: Long Papers)}, pages 16477--16508, Toronto, Canada. Association for Computational Linguistics.

\bibitem[{Geva et~al.(2023)Geva, Bastings, Filippova, and Globerson}]{geva-etal-2023-dissecting}
Mor Geva, Jasmijn Bastings, Katja Filippova, and Amir Globerson. 2023.
\newblock \href {https://doi.org/10.18653/v1/2023.emnlp-main.751} {Dissecting recall of factual associations in auto-regressive language models}.
\newblock In \emph{Proceedings of the 2023 Conference on Empirical Methods in Natural Language Processing}, pages 12216--12235, Singapore. Association for Computational Linguistics.

\bibitem[{Gou et~al.(2023)Gou, Shao, Gong, Shen, Yang, Duan, and Chen}]{gou2023critic}
Zhibin Gou, Zhihong Shao, Yeyun Gong, Yelong Shen, Yujiu Yang, Nan Duan, and Weizhu Chen. 2023.
\newblock \href {http://arxiv.org/abs/2305.11738} {Critic: Large language models can self-correct with tool-interactive critiquing}.

\bibitem[{Hendel et~al.(2023)Hendel, Geva, and Globerson}]{hendel-etal-2023-context}
Roee Hendel, Mor Geva, and Amir Globerson. 2023.
\newblock \href {https://doi.org/10.18653/v1/2023.findings-emnlp.624} {In-context learning creates task vectors}.
\newblock In \emph{Findings of the Association for Computational Linguistics: EMNLP 2023}, pages 9318--9333, Singapore. Association for Computational Linguistics.

\bibitem[{Huang et~al.(2023{\natexlab{a}})Huang, Yu, Ma, Zhong, Feng, Wang, Chen, Peng, Feng, Qin, and Liu}]{huang2023survey_of_hallucination}
Lei Huang, Weijiang Yu, Weitao Ma, Weihong Zhong, Zhangyin Feng, Haotian Wang, Qianglong Chen, Weihua Peng, Xiaocheng Feng, Bing Qin, and Ting Liu. 2023{\natexlab{a}}.
\newblock \href {http://arxiv.org/abs/2311.05232} {A survey on hallucination in large language models: Principles, taxonomy, challenges, and open questions}.

\bibitem[{Huang et~al.(2023{\natexlab{b}})Huang, Song, Wang, Zhao, Chen, Juefei-Xu, and Ma}]{huang2023look-before-you-leap}
Yuheng Huang, Jiayang Song, Zhijie Wang, Shengming Zhao, Huaming Chen, Felix Juefei-Xu, and Lei Ma. 2023{\natexlab{b}}.
\newblock \href {http://arxiv.org/abs/2307.10236} {Look before you leap: An exploratory study of uncertainty measurement for large language models}.

\bibitem[{Jiao et~al.(2023)Jiao, Wang, tse Huang, Wang, Shi, and Tu}]{jiao2023chatgpt-a-good-translator}
Wenxiang Jiao, Wenxuan Wang, Jen tse Huang, Xing Wang, Shuming Shi, and Zhaopeng Tu. 2023.
\newblock \href {http://arxiv.org/abs/2301.08745} {Is chatgpt a good translator? yes with gpt-4 as the engine}.

\bibitem[{Kim et~al.(2023)Kim, Htut, Bowman, and Petty}]{kim-etal-2023-qa}
Najoung Kim, Phu~Mon Htut, Samuel~R. Bowman, and Jackson Petty. 2023.
\newblock \href {https://doi.org/10.18653/v1/2023.acl-long.472} {({QA})$^2$: Question answering with questionable assumptions}.
\newblock In \emph{Proceedings of the 61st Annual Meeting of the Association for Computational Linguistics (Volume 1: Long Papers)}, pages 8466--8487, Toronto, Canada. Association for Computational Linguistics.

\bibitem[{Kuhn et~al.(2023)Kuhn, Gal, and Farquhar}]{semantic-uncertainty-iclr2023}
Lorenz Kuhn, Yarin Gal, and Sebastian Farquhar. 2023.
\newblock \href {https://openreview.net/pdf?id=VD-AYtP0dve} {Semantic uncertainty: Linguistic invariances for uncertainty estimation in natural language generation}.
\newblock In \emph{The Eleventh International Conference on Learning Representations, {ICLR} 2023, Kigali, Rwanda, May 1-5, 2023}. OpenReview.net.

\bibitem[{Lee et~al.(2024)Lee, Bai, Pres, Wattenberg, Kummerfeld, and Mihalcea}]{lee2024mechanistic_alignment}
Andrew Lee, Xiaoyan Bai, Itamar Pres, Martin Wattenberg, Jonathan~K. Kummerfeld, and Rada Mihalcea. 2024.
\newblock \href {http://arxiv.org/abs/2401.01967} {A mechanistic understanding of alignment algorithms: A case study on dpo and toxicity}.

\bibitem[{Li et~al.(2023{\natexlab{a}})Li, Cheng, Zhao, Nie, and Wen}]{li2023halueval}
Junyi Li, Xiaoxue Cheng, Xin Zhao, Jian-Yun Nie, and Ji-Rong Wen. 2023{\natexlab{a}}.
\newblock \href {https://doi.org/10.18653/v1/2023.emnlp-main.397} {{H}alu{E}val: A large-scale hallucination evaluation benchmark for large language models}.
\newblock In \emph{Proceedings of the 2023 Conference on Empirical Methods in Natural Language Processing}, pages 6449--6464, Singapore. Association for Computational Linguistics.

\bibitem[{Li et~al.(2023{\natexlab{b}})Li, Patel, Viégas, Pfister, and Wattenberg}]{li2023inference-time-intervention}
Kenneth Li, Oam Patel, Fernanda Viégas, Hanspeter Pfister, and Martin Wattenberg. 2023{\natexlab{b}}.
\newblock \href {http://arxiv.org/abs/2306.03341} {Inference-time intervention: Eliciting truthful answers from a language model}.

\bibitem[{Manakul et~al.(2023)Manakul, Liusie, and Gales}]{manakul2023selfcheckgpt}
Potsawee Manakul, Adian Liusie, and Mark Gales. 2023.
\newblock \href {https://doi.org/10.18653/v1/2023.emnlp-main.557} {{S}elf{C}heck{GPT}: Zero-resource black-box hallucination detection for generative large language models}.
\newblock In \emph{Proceedings of the 2023 Conference on Empirical Methods in Natural Language Processing}, pages 9004--9017, Singapore. Association for Computational Linguistics.

\bibitem[{Meng et~al.(2022)Meng, Bau, Andonian, and Belinkov}]{meng2022locating-and-editing}
Kevin Meng, David Bau, Alex Andonian, and Yonatan Belinkov. 2022.
\newblock \href {http://papers.nips.cc/paper\_files/paper/2022/hash/6f1d43d5a82a37e89b0665b33bf3a182-Abstract-Conference.html} {Locating and editing factual associations in {GPT}}.
\newblock In \emph{Advances in Neural Information Processing Systems 35: Annual Conference on Neural Information Processing Systems 2022, NeurIPS 2022, New Orleans, LA, USA, November 28 - December 9, 2022}.

\bibitem[{Min et~al.(2023)Min, Krishna, Lyu, Lewis, Yih, Koh, Iyyer, Zettlemoyer, and Hajishirzi}]{min2023factscore}
Sewon Min, Kalpesh Krishna, Xinxi Lyu, Mike Lewis, Wen-tau Yih, Pang Koh, Mohit Iyyer, Luke Zettlemoyer, and Hannaneh Hajishirzi. 2023.
\newblock \href {https://doi.org/10.18653/v1/2023.emnlp-main.741} {{FA}ct{S}core: Fine-grained atomic evaluation of factual precision in long form text generation}.
\newblock In \emph{Proceedings of the 2023 Conference on Empirical Methods in Natural Language Processing}, pages 12076--12100, Singapore. Association for Computational Linguistics.

\bibitem[{Mündler et~al.(2023)Mündler, He, Jenko, and Vechev}]{mündler2023selfcontradictory}
Niels Mündler, Jingxuan He, Slobodan Jenko, and Martin Vechev. 2023.
\newblock \href {http://arxiv.org/abs/2305.15852} {Self-contradictory hallucinations of large language models: Evaluation, detection and mitigation}.

\bibitem[{Ramakrishna et~al.(2023)Ramakrishna, Gupta, Lehmann, and Ziyadi}]{ramakrishna-etal-2023-invite-testbed}
Anil Ramakrishna, Rahul Gupta, Jens Lehmann, and Morteza Ziyadi. 2023.
\newblock \href {https://doi.org/10.18653/v1/2023.findings-emnlp.360} {{INVITE}: a testbed of automatically generated invalid questions to evaluate large language models for hallucinations}.
\newblock In \emph{Findings of the Association for Computational Linguistics: EMNLP 2023}, pages 5422--5429, Singapore. Association for Computational Linguistics.

\bibitem[{Stolfo et~al.(2023)Stolfo, Belinkov, and Sachan}]{stolfo2023mechanistic_reasoning}
Alessandro Stolfo, Yonatan Belinkov, and Mrinmaya Sachan. 2023.
\newblock \href {https://doi.org/10.18653/v1/2023.emnlp-main.435} {A mechanistic interpretation of arithmetic reasoning in language models using causal mediation analysis}.
\newblock In \emph{Proceedings of the 2023 Conference on Empirical Methods in Natural Language Processing}, pages 7035--7052, Singapore. Association for Computational Linguistics.

\bibitem[{Todd et~al.(2023)Todd, Li, Sharma, Mueller, Wallace, and Bau}]{todd2023function_vectors}
Eric Todd, Millicent~L. Li, Arnab~Sen Sharma, Aaron Mueller, Byron~C. Wallace, and David Bau. 2023.
\newblock \href {http://arxiv.org/abs/2310.15213} {Function vectors in large language models}.

\bibitem[{Touvron et~al.(2023)Touvron, Martin, Stone, Albert, Almahairi, Babaei, Bashlykov, Batra, Bhargava, Bhosale, Bikel, Blecher, Ferrer, Chen, Cucurull, Esiobu, Fernandes, Fu, Fu, Fuller, Gao, Goswami, Goyal, Hartshorn, Hosseini, Hou, Inan, Kardas, Kerkez, Khabsa, Kloumann, Korenev, Koura, Lachaux, Lavril, Lee, Liskovich, Lu, Mao, Martinet, Mihaylov, Mishra, Molybog, Nie, Poulton, Reizenstein, Rungta, Saladi, Schelten, Silva, Smith, Subramanian, Tan, Tang, Taylor, Williams, Kuan, Xu, Yan, Zarov, Zhang, Fan, Kambadur, Narang, Rodriguez, Stojnic, Edunov, and Scialom}]{touvron2023llama}
Hugo Touvron, Louis Martin, Kevin Stone, Peter Albert, Amjad Almahairi, Yasmine Babaei, Nikolay Bashlykov, Soumya Batra, Prajjwal Bhargava, Shruti Bhosale, Dan Bikel, Lukas Blecher, Cristian~Canton Ferrer, Moya Chen, Guillem Cucurull, David Esiobu, Jude Fernandes, Jeremy Fu, Wenyin Fu, Brian Fuller, Cynthia Gao, Vedanuj Goswami, Naman Goyal, Anthony Hartshorn, Saghar Hosseini, Rui Hou, Hakan Inan, Marcin Kardas, Viktor Kerkez, Madian Khabsa, Isabel Kloumann, Artem Korenev, Punit~Singh Koura, Marie-Anne Lachaux, Thibaut Lavril, Jenya Lee, Diana Liskovich, Yinghai Lu, Yuning Mao, Xavier Martinet, Todor Mihaylov, Pushkar Mishra, Igor Molybog, Yixin Nie, Andrew Poulton, Jeremy Reizenstein, Rashi Rungta, Kalyan Saladi, Alan Schelten, Ruan Silva, Eric~Michael Smith, Ranjan Subramanian, Xiaoqing~Ellen Tan, Binh Tang, Ross Taylor, Adina Williams, Jian~Xiang Kuan, Puxin Xu, Zheng Yan, Iliyan Zarov, Yuchen Zhang, Angela Fan, Melanie Kambadur, Sharan Narang, Aurelien Rodriguez, Robert Stojnic, Sergey Edunov, and Thomas
  Scialom. 2023.
\newblock \href {http://arxiv.org/abs/2307.09288} {Llama 2: Open foundation and fine-tuned chat models}.

\bibitem[{Trivedi et~al.(2023)Trivedi, Balasubramanian, Khot, and Sabharwal}]{trivedi-etal-2023-interleaving-retrieval}
Harsh Trivedi, Niranjan Balasubramanian, Tushar Khot, and Ashish Sabharwal. 2023.
\newblock \href {https://doi.org/10.18653/v1/2023.acl-long.557} {Interleaving retrieval with chain-of-thought reasoning for knowledge-intensive multi-step questions}.
\newblock In \emph{Proceedings of the 61st Annual Meeting of the Association for Computational Linguistics (Volume 1: Long Papers)}, pages 10014--10037, Toronto, Canada. Association for Computational Linguistics.

\bibitem[{Vaswani et~al.(2017)Vaswani, Shazeer, Parmar, Uszkoreit, Jones, Gomez, Kaiser, and Polosukhin}]{vaswani2017attention-is-all-you-need}
Ashish Vaswani, Noam Shazeer, Niki Parmar, Jakob Uszkoreit, Llion Jones, Aidan~N. Gomez, Lukasz Kaiser, and Illia Polosukhin. 2017.
\newblock \href {https://proceedings.neurips.cc/paper/2017/hash/3f5ee243547dee91fbd053c1c4a845aa-Abstract.html} {Attention is all you need}.
\newblock In \emph{Advances in Neural Information Processing Systems 30: Annual Conference on Neural Information Processing Systems 2017, December 4-9, 2017, Long Beach, CA, {USA}}, pages 5998--6008.

\bibitem[{Vu et~al.(2023)Vu, Iyyer, Wang, Constant, Wei, Wei, Tar, Sung, Zhou, Le, and Luong}]{vu2023freshllms}
Tu~Vu, Mohit Iyyer, Xuezhi Wang, Noah Constant, Jerry Wei, Jason Wei, Chris Tar, Yun-Hsuan Sung, Denny Zhou, Quoc Le, and Thang Luong. 2023.
\newblock \href {http://arxiv.org/abs/2310.03214} {Freshllms: Refreshing large language models with search engine augmentation}.

\bibitem[{Wang et~al.(2023)Wang, Li, Dai, Chen, Zhou, Meng, Zhou, and Sun}]{wang-etal-2023-label-words-are-anchors}
Lean Wang, Lei Li, Damai Dai, Deli Chen, Hao Zhou, Fandong Meng, Jie Zhou, and Xu~Sun. 2023.
\newblock \href {https://doi.org/10.18653/v1/2023.emnlp-main.609} {Label words are anchors: An information flow perspective for understanding in-context learning}.
\newblock In \emph{Proceedings of the 2023 Conference on Empirical Methods in Natural Language Processing}, pages 9840--9855, Singapore. Association for Computational Linguistics.

\bibitem[{Wei et~al.(2022)Wei, Wang, Schuurmans, Bosma, Ichter, Xia, Chi, Le, and Zhou}]{wei2023chainofthought}
Jason Wei, Xuezhi Wang, Dale Schuurmans, Maarten Bosma, Brian Ichter, Fei Xia, Ed~H. Chi, Quoc~V. Le, and Denny Zhou. 2022.
\newblock \href {http://papers.nips.cc/paper\_files/paper/2022/hash/9d5609613524ecf4f15af0f7b31abca4-Abstract-Conference.html} {Chain-of-thought prompting elicits reasoning in large language models}.
\newblock In \emph{Advances in Neural Information Processing Systems 35: Annual Conference on Neural Information Processing Systems 2022, NeurIPS 2022, New Orleans, LA, USA, November 28 - December 9, 2022}.

\bibitem[{Xu et~al.(2023)Xu, Hong, Li, Hu, Chen, and Zhang}]{xu2023tool-manipulation-capability}
Qiantong Xu, Fenglu Hong, Bo~Li, Changran Hu, Zhengyu Chen, and Jian Zhang. 2023.
\newblock \href {http://arxiv.org/abs/2305.16504} {On the tool manipulation capability of open-source large language models}.

\bibitem[{Yang et~al.(2023)Yang, Sun, and Wan}]{yang-etal-2023-new-benchmark-phd}
Shiping Yang, Renliang Sun, and Xiaojun Wan. 2023.
\newblock \href {https://doi.org/10.18653/v1/2023.findings-emnlp.256} {A new benchmark and reverse validation method for passage-level hallucination detection}.
\newblock In \emph{Findings of the Association for Computational Linguistics: EMNLP 2023}, pages 3898--3908, Singapore. Association for Computational Linguistics.

\bibitem[{Yu et~al.(2023{\natexlab{a}})Yu, Merullo, and Pavlick}]{yu-etal-2023-characterizing}
Qinan Yu, Jack Merullo, and Ellie Pavlick. 2023{\natexlab{a}}.
\newblock \href {https://doi.org/10.18653/v1/2023.emnlp-main.615} {Characterizing mechanisms for factual recall in language models}.
\newblock In \emph{Proceedings of the 2023 Conference on Empirical Methods in Natural Language Processing}, pages 9924--9959, Singapore. Association for Computational Linguistics.

\bibitem[{Yu et~al.(2023{\natexlab{b}})Yu, Min, Zettlemoyer, and Hajishirzi}]{yu-etal-2023-crepe}
Xinyan Yu, Sewon Min, Luke Zettlemoyer, and Hannaneh Hajishirzi. 2023{\natexlab{b}}.
\newblock \href {https://doi.org/10.18653/v1/2023.acl-long.583} {{CREPE}: Open-domain question answering with false presuppositions}.
\newblock In \emph{Proceedings of the 61st Annual Meeting of the Association for Computational Linguistics (Volume 1: Long Papers)}, pages 10457--10480, Toronto, Canada. Association for Computational Linguistics.

\bibitem[{Yuksekgonul et~al.(2023)Yuksekgonul, Chandrasekaran, Jones, Gunasekar, Naik, Palangi, Kamar, and Nushi}]{yuksekgonul2023attention-satisfies}
Mert Yuksekgonul, Varun Chandrasekaran, Erik Jones, Suriya Gunasekar, Ranjita Naik, Hamid Palangi, Ece Kamar, and Besmira Nushi. 2023.
\newblock \href {http://arxiv.org/abs/2309.15098} {Attention satisfies: A constraint-satisfaction lens on factual errors of language models}.

\bibitem[{Zhang et~al.(2023)Zhang, Li, Cui, Cai, Liu, Fu, Huang, Zhao, Zhang, Chen, Wang, Luu, Bi, Shi, and Shi}]{zhang2023sirens_song}
Yue Zhang, Yafu Li, Leyang Cui, Deng Cai, Lemao Liu, Tingchen Fu, Xinting Huang, Enbo Zhao, Yu~Zhang, Yulong Chen, Longyue Wang, Anh~Tuan Luu, Wei Bi, Freda Shi, and Shuming Shi. 2023.
\newblock \href {http://arxiv.org/abs/2309.01219} {Siren's song in the ai ocean: A survey on hallucination in large language models}.

\bibitem[{Zhao et~al.(2024)Zhao, Chen, Yang, Liu, Deng, Cai, Wang, Yin, and Du}]{ExplainabilityForLLMSurvey}
Haiyan Zhao, Hanjie Chen, Fan Yang, Ninghao Liu, Huiqi Deng, Hengyi Cai, Shuaiqiang Wang, Dawei Yin, and Mengnan Du. 2024.
\newblock \href {https://doi.org/10.1145/3639372} {Explainability for large language models: A survey}.
\newblock \emph{ACM Trans. Intell. Syst. Technol.}
\newblock Just Accepted.

\bibitem[{Zhou et~al.(2023)Zhou, Cui, Yoon, Zhang, Deng, Finn, Bansal, and Yao}]{zhou2023analyzing}
Yiyang Zhou, Chenhang Cui, Jaehong Yoon, Linjun Zhang, Zhun Deng, Chelsea Finn, Mohit Bansal, and Huaxiu Yao. 2023.
\newblock \href {http://arxiv.org/abs/2310.00754} {Analyzing and mitigating object hallucination in large vision-language models}.

\bibitem[{Zou et~al.(2023)Zou, Phan, Chen, Campbell, Guo, Ren, Pan, Yin, Mazeika, Dombrowski, Goel, Li, Byun, Wang, Mallen, Basart, Koyejo, Song, Fredrikson, Kolter, and Hendrycks}]{zou2023transparency}
Andy Zou, Long Phan, Sarah Chen, James Campbell, Phillip Guo, Richard Ren, Alexander Pan, Xuwang Yin, Mantas Mazeika, Ann-Kathrin Dombrowski, Shashwat Goel, Nathaniel Li, Michael~J. Byun, Zifan Wang, Alex Mallen, Steven Basart, Sanmi Koyejo, Dawn Song, Matt Fredrikson, Zico Kolter, and Dan Hendrycks. 2023.
\newblock \href {http://arxiv.org/abs/2310.01405} {Representation engineering: A top-down approach to ai transparency}.

\end{thebibliography}

\clearpage

\appendix

\begin{table*}[b!]
\resizebox{\linewidth}{!}{
\begin{tabular}{c|c|cl}
\hline
    \arrayrulecolor{gray!100}
\multicolumn{1}{c|}{Datasets} & Corrupted Triple                                      & \multicolumn{2}{c}{Question Templates}                                                                                \\ \hline
\multirow{4}{*}{Prize}       & \multirow{4}{*}{\begin{tabular}[c]{@{}c@{}}(person, \\ was awarded, \\ false prize)\end{tabular}}     & T1 & For what specific contribution was <person> awarded  <false prize>?  \\
                             &                                                       & T2 & <person> was awarded <false prize> for what specific reason?        \\
                             &                                                       & T3 & <false prize> was awarded to <person>  for what specific reason?       \\
                             &                                                       & T4 & Why was <person> awarded the <false prize>?                         \\ \hline
\multirow{4}{*}{Movie}       & \multirow{4}{*}{\begin{tabular}[c]{@{}c@{}}(movie, \\ was released in,\\  false time)\end{tabular}}  & T1 & Why was the film <movie> released in <false time>{}?                   \\
                             &                                                       & T2 & What was the film <movie> released in <false time>   describing about? \\
                             &                                                       & T3 & What was the <false time> film <movie> about?                          \\
                             &                                                       & T4 & Who are the main characters in the <false time> film   <movie>?     \\     \arrayrulecolor{black} \hline
\end{tabular}
}
\caption{Details of the question templates in our datasets.}
\label{question_templates_table}
\end{table*}

\section{Question Templates}
\label{appendix:question_templates}

Details of the concrete question templates we employed are provided in Table \ref{question_templates_table}.

\section{Results of Generalization Experiments using model Llama-2-13b-chat}
\label{appendix:generalization}

Table \ref{generalization_13b} shows the experimental results of the generalization experiments using model Llama-2-13b-chat.

\begin{table}[hb]
\resizebox{\linewidth}{!}{
\begin{tabular}{cr|ccccc} \hline
    \arrayrulecolor{gray!100}
\multicolumn{2}{c|}{Methods}                              & T1    & T2    & T3    & T4    & Avg   \\ \hline
\multicolumn{2}{c|}{FAITH}                                & 82.02 & 80.52 & 49.65 & 23.78 & \textbf{58.99 }\\  
\rowcolor[HTML]{E7E6E6} 
\cellcolor[HTML]{E7E6E6}                         & w/T1  & 82.02 & 84.52 & 20.28 & 13.19 & 50.00 \\
\rowcolor[HTML]{E7E6E6} 
\cellcolor[HTML]{E7E6E6}                         & w/T2  & 77.92 & 80.52 & 11.69 & 10.19 & 45.08 \\
\rowcolor[HTML]{E7E6E6} 
\cellcolor[HTML]{E7E6E6}                         & w/T3  & 80.72 & 73.33 & 49.65 & 46.45 & \textbf{62.54} \\
\rowcolor[HTML]{E7E6E6} 
\multirow{-4}{*}{\cellcolor[HTML]{E7E6E6}Within} & w/T4  & 81.52 & 81.92 & 28.77 & 23.78 & 54.00 \\  
                                                 & w/PT1 & 68.33 & 63.94 & 12.09 & 8.59  & 38.24 \\
                                                 & w/PT2 & 79.42 & 77.12 & 2.70   & 2.10  & 40.34 \\
                                                 & w/PT3 & 81.72 & 80.82 & 5.00     & 3.70   & 42.81 \\
\multirow{-4}{*}{Across}                         & w/PT4 & 80.52 & 77.92 & 51.25 & 42.36 & \textbf{63.01} \\ \hline
\rowcolor[HTML]{E7E6E6} 
\multicolumn{2}{c|}{Random}   & 50.55 & 41.86 & 2.40  & 0.70  & \textbf{23.88 } \\ 
    \arrayrulecolor{black}
\hline
\end{tabular}
}
\caption{Generalizability of the attention heads on the 13B model. ``w/T1'' denotes using the false premise heads identified on the question template 1 in the same Movie dataset. ``w/PT1"" denotes using the false premise heads identified on the question template 1 in the Prize dataset. }
\label{generalization_13b}
\end{table}

\end{document}